\documentclass[10pt,twocolumn,letterpaper]{article}

\usepackage{cvpr}              %

\usepackage[dvipsnames]{xcolor}

\definecolor{cvprblue}{rgb}{0.21,0.49,0.74}
\usepackage[pagebackref,breaklinks,colorlinks,citecolor=cvprblue]{hyperref}

\usepackage{lipsum}
\usepackage[accsupp]{axessibility} %

\newcommand\blfootnote[1]{%
  \begingroup
  \renewcommand\thefootnote{}\footnote{#1}%
  \addtocounter{footnote}{-1}%
  \endgroup
}

\newcommand{\gptfv}{GPT4V}
\newcommand{\benchmark}{DESIGNERINTENTION}
\newcommand{\deepfloyd}{DeepFloydIF}
\newcommand{\sdxl}{SDXL1.0}
\newcommand{\dalle}{DALL-E3}
\newcommand{\llava}{LLaVA}

\def\paperID{3} %
\def\confName{CVPR}
\def\confYear{2024}

\title{OpenCOLE: Towards Reproducible Automatic Graphic Design Generation}

\author{
    Naoto Inoue\textsuperscript{*} \quad Kento Masui\textsuperscript{*} \quad Wataru Shimoda\textsuperscript{*} \quad Kota Yamaguchi \\
    CyberAgent
}

\begin{document}
\twocolumn[{%
\renewcommand\twocolumn[1][]{#1}%
\maketitle
\begin{center}
    \centering
    \captionsetup{type=figure}
    \includegraphics[width=\textwidth]{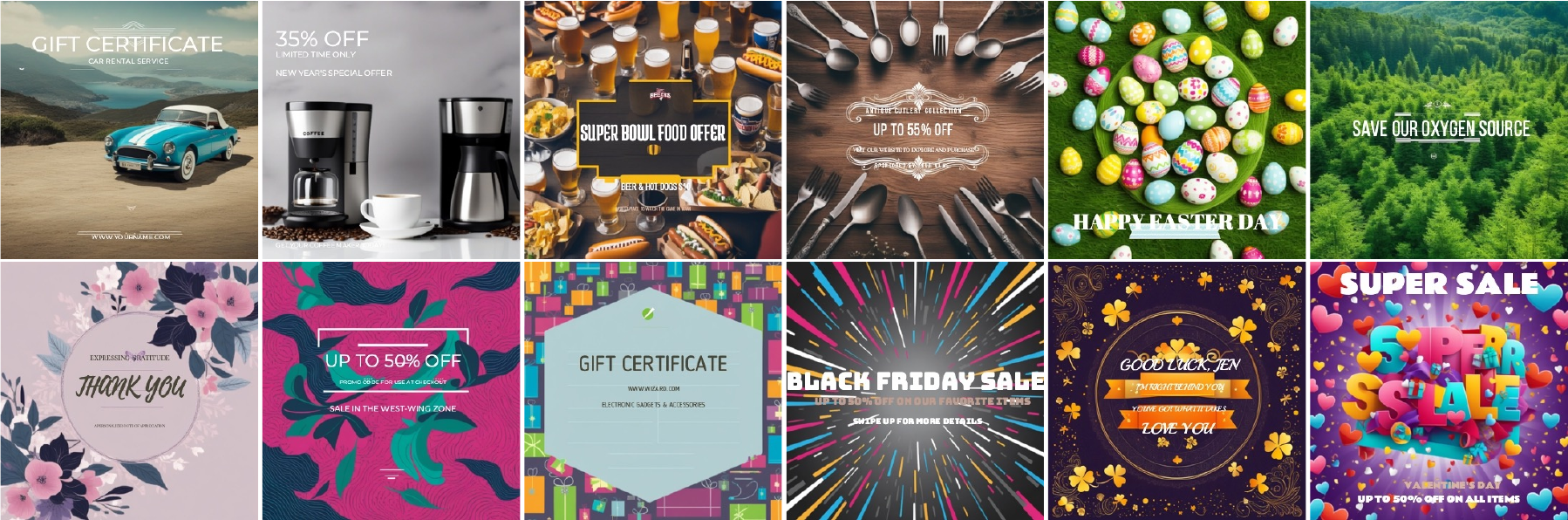}
    \caption{Generated graphic designs from OpenCOLE, given short user intentions.}
    \label{fig:examples}
\end{center}%
}]

\blfootnote{\textsuperscript{*}: equal contribution}

\begin{abstract}
Automatic generation of graphic designs has recently received considerable attention.
However, the state-of-the-art approaches are complex and rely on proprietary datasets, which creates reproducibility barriers.
In this paper, we propose an open framework for automatic graphic design called OpenCOLE, where we build a modified version of the pioneering COLE and train our model exclusively on publicly available datasets.
Based on GPT4V evaluations, our model shows promising performance comparable to the original COLE.
We release the pipeline and training results to encourage open development.\footnote{\url{https://github.com/CyberAgentAILab/OpenCOLE}}
\end{abstract}

\section{Introduction}

Graphic design serves as an essential medium for visual communication and is ubiquitously present in our daily lives. It encompasses a variety of multi-modal elements such as text or images, and the composition of these elements necessitates an in-depth comprehension of many aesthetic facets, including layout and readability.

Automatic graphic design has long been a goal within the research community, with a rich history of exploration and experimentation~\cite{lok2001survey,agrawala2011design,yang2016automatic}. Over the past decade, learning-based formulations, particularly those employing neural networks, have been utilized in various approaches aimed at resolving specific design sub-tasks, such as layout generation~\cite{li2019layoutgan,jyothi2019layoutvae,gupta2021layout,Kikuchi2021,inoue2023layout,zhang_layoutdiffusion,layoutFormerPP}, font recommendation~\cite{zhao2018modeling,fontpair,shimoda2024towards}, and colorization~\cite{yuan2021infocolorizer,qiu2023color,qiu2023multimodal,kikuchi2023generative,shi2023stijl}.  %
With the substantial advancements in modeling texts and images, some ambitious papers have begun to tackle the generation or editing of complete graphic designs~\cite{yamaguchi2021canvasvae,inoue2023towards,lin2023autoposter}.

Most recently, COLE~\cite{jia2023cole} has exhibited outstanding proficiency in generating graphics based on brief intention prompts.
COLE deconstructs the intricate generation procedure into several distinct stages.
The outputs comprise a background layer, an object image layer, and text layers with a broad spectrum of typographic properties such as font type and line spacing.
An off-the-shelf graphic renderer rasterizes these layers to obtain the preview image.
Owing to this methodological breakdown, texts are presumed to maintain legibility and editability, which is a crucial requirement in graphic design.

The primary shortcoming of COLE is its lack of publicly accessible datasets and codes.
Because of its cascaded process and reliance on a graphical renderer, its proprietary nature poses significant challenges to reproduction.
We propose OpenCOLE, an open-source implementation of COLE that utilizes only publicly available datasets and models.
This aims to democratize the development of generative models for graphic design and foster a more inclusive community.
Following COLE, we present a framework composed of three stages with minor modifications.
For the dataset, we utilize Crello~\cite{yamaguchi2021canvasvae}, which is a publicly available graphic design dataset, and synthesize all the necessary data for training each stage of OpenCOLE.

We experimentally show that our OpenCOLE matches COLE in terms of GPT4V-based evaluation on five aspects specific to graphic design on \benchmark{} benchmark~\cite{jia2023cole}. We also compare OpenCOLE with LLM-augmented text-to-image models and show it performs better than \deepfloyd{}~\cite{deepfloyd_if} but it still lags behind \sdxl{}~\cite{podell2023sdxl} and \dalle{}~\cite{dall_e3}.
We further discuss potential challenges to promote the research for automatic graphic design generation.

Our contributions can be summarized as follows:
\begin{itemize}[noitemsep,nolistsep,leftmargin=*]
\item We propose to construct an open-sourced automatic graphic design generation pipeline by leveraging open-sourced data and models to facilitate the democratization of development.
\item Our empirical results suggest that our OpenCOLE performs almost on par with COLE.
\end{itemize}

\section{Method}

\begin{figure*}[t]
\centering
\includegraphics[width=\hsize]{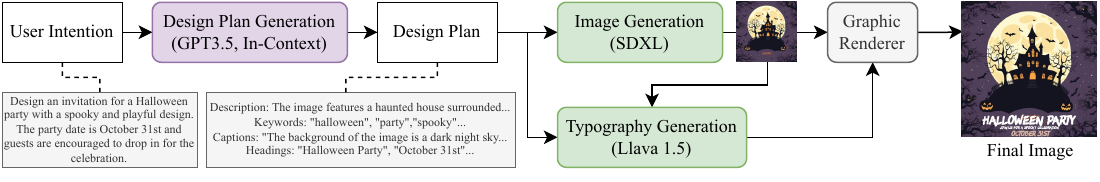}
\caption{An architecture of OpenCOLE. We mimic the architecture of COLE with adjustments. The user intention is first converted to a design plan with GPT3.5 and in-context learning. Then, the image generation module and the typography generation module synthesize design elements following a design plan. Finally, the graphic renderer composes the final image.}
\label{fig:architecture}
\end{figure*}

Following COLE~\cite{jia2023cole}, the aim of our OpenCOLE architecture is to convert a user's intention to create a graphic design into a complete design composed of visual elements.
Our OpenCOLE implementation consists of a design plan generation module, an image generation module, and a typography generation module, as shown in \cref{fig:architecture}.

First, a design plan generation module takes a user's intention and translates it into a detailed explanation and information for design composition in JSON format, which we refer to as a design plan. 
Next, an image generation module generates an image that covers the entire canvas following the design plan.
Then, a typography generation module synthesizes typographic attributes, such as font or color, from the generated image and the design plan.
Finally, a graphic renderer composes the image and the typography attributes to place the texts to form a final design image.

This design choice mimics the architecture of COLE, where the original architecture consists of several corresponding components, namely a Design-LLM, Image Generation Modules, and a Typography-LMM. 
We introduce slight adjustments for each component to complement the ambiguous part in the original COLE paper for our OpenCOLE implementation.

\subsection{Design Plan Generation}

In the design plan generation module, we use an LLM to convert a rather vague user intention into a design plan that works as a schematic of the design composition. This design plan instructs the other modules to generate design elements for the final composition. 

The design plan consists of 4 items: description, keywords, captions, and headings. Detailed explanations are as follows:

\begin{itemize}[noitemsep,nolistsep,leftmargin=*]
    \item Description: A detailed caption of the final design composition. The caption explains what the image is about, the depicted object, and the context. 
    \item Keywords: A list of keywords to express the image in various aspects, such as color, object, and genre.
    \item Captions: A background caption and an object caption. The background caption mainly explains the color and objects behind the main subject. The objects caption explains individual objects that should appear in the design with their coordination.
    \item Headings: A heading and a subheading for the final composed design. Since a design is often made up of multiple texts, we explicitly specify the text here for the typography generation module.
\end{itemize}

We have used a pre-trained LLM with in-context learning prompt~\cite{brown2020incontext} to generate a design plan. The in-context learning prompt contains examples of 5 pairs for a user intention and a design plan to facilitate the generation of a design plan. This approach is a simplification of the original COLE that fine-tunes Llama~\cite{touvron2023llama} for JSON generation.

\subsection{Image Generation}
In the image generation module, we transform a design plan into a graphic design image without text layers.
Since an off-the-shelf text-to-image model tends to yield illegible texts when instructed to generate graphic design images, we fine-tune the generator on a dataset of text-image pairs.
We concatenate the background and object caption in the design plan to form the input text.
We render the original graphic document data, excluding text layers, to create the target image.

This single-stage generation process deviates from the original COLE's two-stage generation. 
COLE first employs a text-to-background model to generate an embellished background image.
Then, COLE uses a text-to-object model to take the background image as an additional conditioning input.
We adopted the single-stage approach due to the challenges of obtaining reliable input-target pairs for both stages.

\subsection{Typography Generation}
In the typography generation module, we fine-tune large multi-modal models (LMMs) to generate typographic attributes given the outputs from the former modules: the design plan and the generated image.
This module follows the original TypographyLMM of COLE.
The typographic attributes consist of fine-grained text styling parameters, such as font, font size, color, and letter spacing.
The typography LMM handles fine-grained attributes as language via JSON format, which contains all typographic attributes in a design for per-text elements with hierarchical structures.

\section{Experiments}
\subsection{Implementation Details}
For the base dataset, we use the Crello dataset~\cite{yamaguchi2021canvasvae}. It provides around 22k design templates for many domains, such as social media posts, banner ads, blog headers, and printed posters. Each sample is stored in a vector format and contains images, texts, layouts, and typography information.

\paragraph{Dataset for In-Context Learning}
For the design plan generation with an LLM and in-context learning, we prepared an intention-to-design-plan dataset following the original COLE with Crello dataset. 

For intention creation, we used GPT3.5 with adjusted prompt from COLE. This adjustment includes adding intention examples that were referenced but not in the original COLE's prompt, and providing information such as title, format, keywords, and texts.

For design plan creation, we followed COLE to use ~\llava{}-1.5-13B model~\cite{liu2023llava}, but with a slightly simplified JSON structure. Since the prompts used for creating a design plan with \llava{} in the original COLE are not available, we implemented our design plan generation using the divide-and-conquer approach.
For each item of a design plan (description, keywords, captions, and headings), we designed a specific prompt to extract only the corresponding information from an image using \llava{}. This is to reduce the complexity of the task and to improve the stability of the LLM in successfully extracting meaningful information as JSON text. After each piece of information for an image is extracted, we merge them to form a single design plan.

\paragraph{Design Plan Generation}
We used GPT3.5 as a base LLM for design plan generation with in-context learning.
We randomly selected 5 examples from the intention-to-design-plan dataset synthesized from the Crello dataset for in-context examples.

\paragraph{Image Generation}
We fine-tune \sdxl{}~\cite{podell2023sdxl} for 10,000 iterations with a constant learning rate of $8 \times 10^{-7}$ with 1,000 warmup steps using Simpletuner~\cite{simpletuner}.
Images with less than 0.5M pixels in the area are filtered to avoid upscaling artifacts.
\sdxl{} has a maximum token length limit of 77. During training, we randomly omit some sentences in the input to meet the limit. During sampling, we employ Compel~\cite{compel} to feed all the information in the text input and generate images with size of $1024\times1024$.

\paragraph{Typography Generation}
Unlike COLE, we merge headings and subheadings into a single list because the Crello dataset does not have such attributes. 
Similarly to COLE, we employ \llava{}1.5-7B for the base LMM. We fine-tune the model for 6 epochs with a batch size of 32, learning rate of 2e-4, max token length of 4096, LoRA rank 128, LoRA alpha 256, and input resolution of $336\times336$.

\subsection{Benchmark, Metrics, and Baselines}
Following COLE~\cite{jia2023cole}, we test all the models on \benchmark{} benchmark, which contains 200 graphic design intention prompts. These prompts span six salient categories within graphic design: advertising, events, marketing, posts, covers \& headers, and creative. The quality of the generated design images is evaluated on a scale from 1 (poor) to 10 (excellent) using \gptfv{}~\cite{gpt4v} to assess five crucial aspects.

We obtain COLE's results from their project page.
Following COLE, we manually test text-to-image models as additional baselines.
These baselines take GPT4-augmented prompts as inputs and use \deepfloyd{}~\cite{deepfloyd_if}, \sdxl{}~\cite{podell2023sdxl}, or \dalle{}~\cite{dall_e3} to generate the final image. 

\subsection{Quantitative Evaluation}
In \cref{tab:quantitative_evaluation}, we see that OpenCOLE performs competitively with the original COLE. For reference, we also show the results of simple text-to-image baselines \benchmark{}. There is still room for improvement for OpenCOLE since it largely lags behind \dalle{} and \sdxl{} in all the aspects, though the texts in the baselines are not editable and sometimes illegible.

{
\setlength{\tabcolsep}{4pt}
\begin{table}[t]
\begin{center}
\caption{Comparison of different models on full \benchmark{}.  \gptfv{} evaluation considers the following aspects:
(i)~\textit{design and layout}, (ii)~\textit{content relevance}, (iii)~\textit{typography and color}, (vi)~\textit{graphics and images}, and (v)~\textit{innovation}. $^{\dag}$ indicates that GPT-4 converts user intentions into detailed prompts fed to text-to-image models. Higher is better.}
\label{tab:quantitative_evaluation}
\begin{tabular}{ccccccc}
     & (i) & (ii) & (iii) & (iv) & (v) & Avg. \\ \toprule
    \multicolumn{7}{l}{{\color{gray} Text fixed}} \\
    DeepFloyd/IF~\cite{deepfloyd_if}$^{\dag}$ & 5.6 & 6.3 & 4.8 & 6.5 & 5.0 & 5.6 \\
    SDXL~\cite{podell2023sdxl}$^{\dag}$ & 7.2 & 7.6 & 7.1 & 7.9 & 6.3 & 7.2 \\ 
    DALL-E3~\cite{dall_e3}$^{\dag}$ & \textbf{7.8} & \textbf{8.3} & \textbf{7.6} & \textbf{8.7} & \textbf{7.2} & \textbf{7.9} \\ \midrule
    \multicolumn{7}{l}{{\color{gray} Text editable}} \\
    COLE~\cite{jia2023cole} & 6.0 & 6.9 & \textbf{5.7} & 6.2 & 5.1 & 6.0 \\
    OpenCOLE & \textbf{6.3} & \textbf{7.0} & 5.6 & \textbf{7.1} & \textbf{5.3} & \textbf{6.3} \\ \bottomrule %
\end{tabular}
\end{center}
\end{table}
}

\subsection{Qualitative Evaluation}
Fig.~\ref{fig:examples} demonstrates the generated designs by OpenCOLE.
We observe that OpenCOLE can generate plausible graphic designs. 
The generated texts do not always accurately reflect intentions, but most of them are at least partially related.
Notably, the generated images tend to contain empty spaces for text placement, and the typography LMM appears to position texts in these spaces by recognizing them.

\begin{figure}[t]
\centering
\includegraphics[width=\hsize]{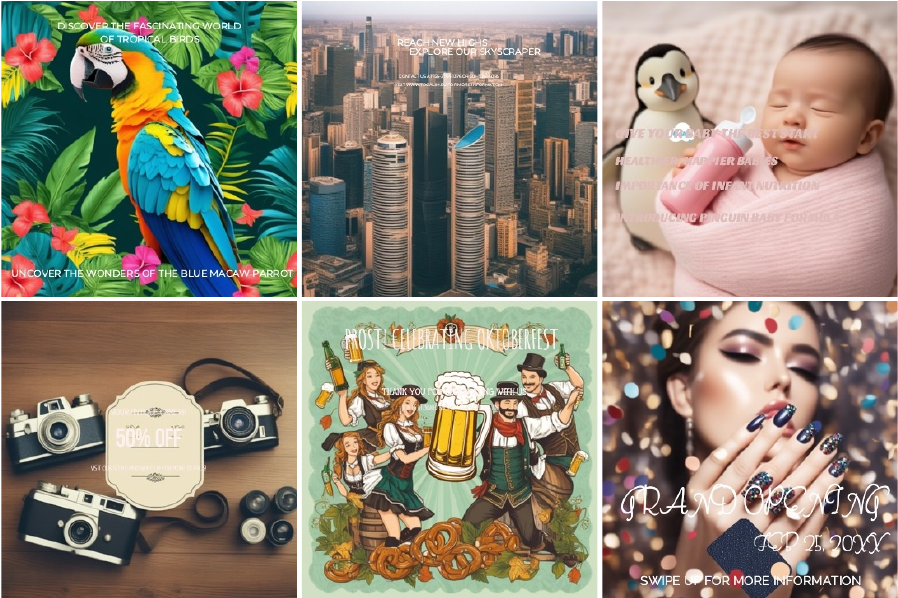}
\caption{Failure cases of OpenCOLE.}
\label{fig:failure_cases}
\end{figure}

Fig.~\ref{fig:failure_cases} shows failure cases of OpenCOLE.
We observe that OpenCOLE tends to generate some low-legibility texts.
Some factors, we believe, include generating excessively long sentences without line breaks, generating images with small empty spaces for text placement, overlooking color contrasts, and choosing very thin fonts.

\begin{figure}[t]
\centering
\includegraphics[width=\hsize]{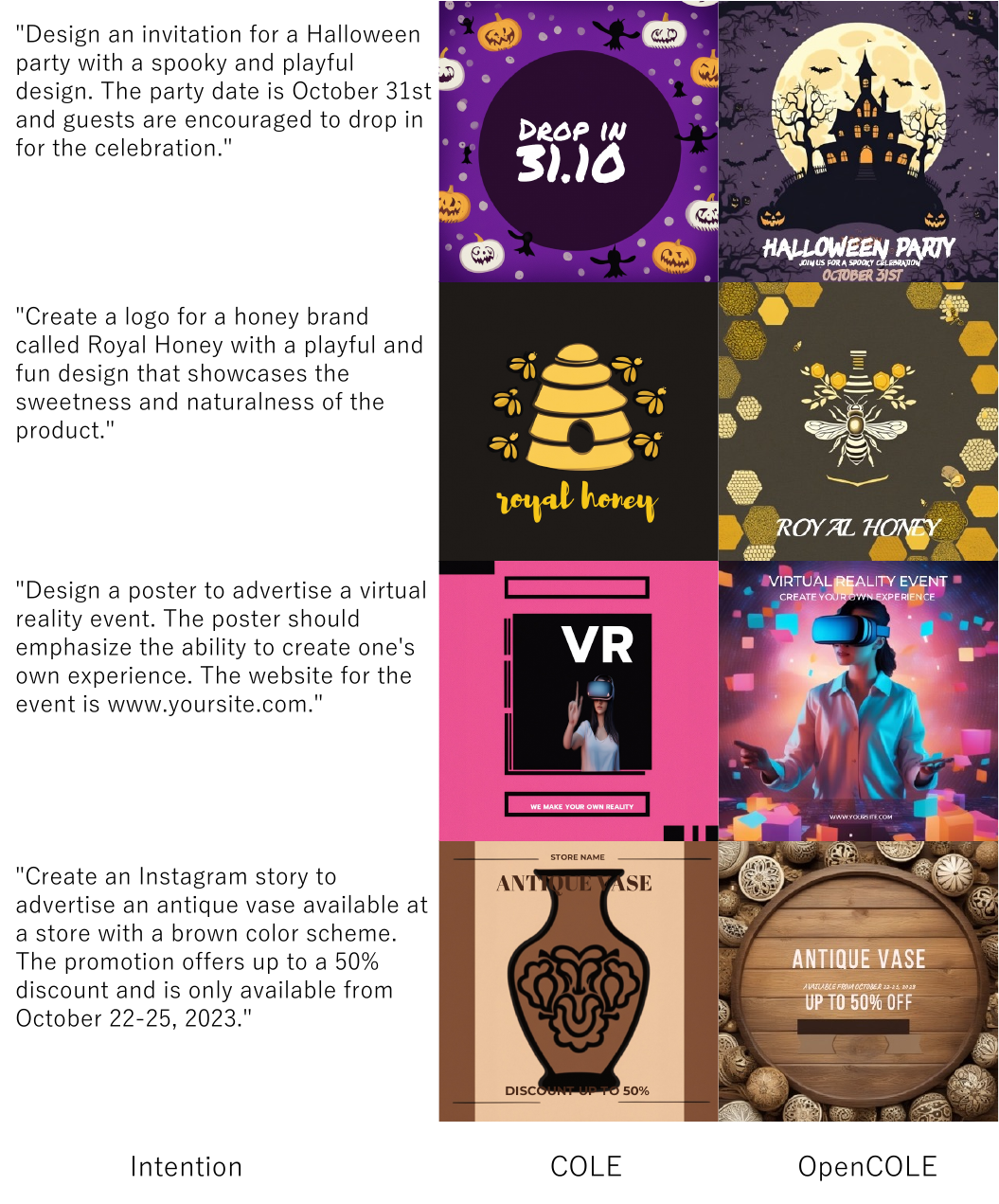}
\caption{The comparisons between COLE and OpenCOLE. The middle and right images are generated designs from the left intentions by COLE and OpenCOLE, respectively. }
\label{fig:comparisons}
\end{figure}

Fig.~\ref{fig:comparisons} presents a comparison between COLE and OpenCOLE.
Both methods can generate plausible designs, but they exhibit different tendencies in designs from them.
While COLE generates simple designs with distinct texts, OpenCOLE generates more informative images with complex texts.
We attribute these differences to variations in the datasets used and modifications made to certain modules.

\section{Discussion}
While we believe that OpenCOLE constitutes a pivotal initial stride towards the open-source progression of automated graphic design generation, many discussions persist. The most notable point is the dependency on black box \gptfv{} assessment. We observe that some generated images without textual content are highly ranked by \gptfv{}. This often happens in GPT4-augmented text-to-image baselines. Evaluation considering both the input intention and output-generated images would be necessary.

{
    \small
    \bibliographystyle{ieeenat_fullname}
    \bibliography{main}
}

\end{document}